\documentclass[namedrefereces]{kluwer}
\usepackage{graphicx}
\renewcommand{\baselinestretch}{\currentspacing}

\newcommand{\symbolpar}{3.5in}
\newcommand{\symbolsep}{& $\bullet$ & \parbox[t]{\symbolpar}}
\newcommand{\logand}{\wedge}
\newcommand{\logor}{\vee}
\newcommand{\Logand}{\bigwedge}
\begin{document}
\begin{article}
% input the header including the title and abstract.
%\def\BibTeX{{\rm B\kern-.05em{\sc i\kern-.025em b}\kern-.08em
%    T\kern-.1667em\lower.7ex\hbox{E}\kern-.125emX}}
\begin{opening}
\title{Fuzzy Logic Classification of Imaging Laser
Desorption Fourier Transform Mass Spectrometry Data }
\author{Timothy R. \surname{McJunkin}
\email{timothy.mcjunkin@inl.gov}}
%\thanks{
%    This work was supported by LDRD program for the United States Department of
%    Energy, under contract DE-AC07-99ID13727
%  }}
\institute{ Idaho National Laboratory, P.O. Box 1625, Idaho Falls,
ID
  83415-2210}
\author{Jill R. \surname{Scott} \email{jill.scott@inl.gov}}
\institute{ Idaho National Laboratory, P.O. Box 1625, Idaho Falls,
ID 83415-2208}

\runningauthor{T.R. McJunkin and J.R. Scott} \runningtitle{Fuzzy
Logic Classification of Imaging Laser Desorption FTMS}

%\markboth{Fuzzy Optimization and Decision Making}
%{Murray and Balemi: Using the style file IEEEtran.sty} %!PN
%{
% T. R. McJunkin, J. R. Scott, and P. L. Tremblay:
%Fuzzy Logic Classification of Basalt Minerals with Laser
%Desorption Fourier Transform Mass Spectrometer Data
%
%} %!PN

%\pubid{0000--0000/00\$00.00~US Government Work Not Protected by
%US Copyright}

%\maketitle
%\thispagestyle{plain}\pagestyle{plain}

\begin{abstract}
  A fuzzy logic based classification engine has been developed for
classifying mass spectra obtained with an imaging internal source
Fourier transform mass spectrometer (I$^{2}$LD-FTMS).
Traditionally, an operator uses the relative abundance of ions
with specific mass-to-charge (m/z) ratios to categorize spectra.
An operator does this by comparing the spectrum of m/z versus
abundance of an unknown sample against a library of spectra from
known samples. Automated positioning and acquisition allow
I$^{2}$LD-FTMS to acquire data from very large grids, this would
require classification of up to 3600 spectrum per hour to keep
pace with the acquisition. The tedious job of classifying numerous
spectra generated in an I$^{2}$LD-FTMS imaging application can be
replaced by a fuzzy rule base if the cues an operator uses can be
encapsulated. We present the translation of linguistic rules to a
fuzzy classifier for mineral phases in basalt. This paper also
describes a method for gathering statistics on ions, which are not
currently used in the rule base, but which may be candidates for
making the rule base more accurate and complete or to form new
rule bases based on data obtained from known samples. A spatial
method for classifying spectra with low membership values, based
on neighboring sample classifications, is also presented.

\end{abstract}


\end{opening}

%\input{header_elsart.tex}
%-\thispagestyle{empty}

%\fedsection{Introduction} \label{intro_sec}
\section{Introduction} \label{intro_sec}
The Idaho National Laboratory (INL) has produced an imaging
internal laser desorption Fourier transform mass spectrometer
(I$^{2}$LD-FTMS) that provides the chemical imaging for the
laser-based optical and chemical imager (LOCI). The I$^{2}$LD-FTMS
couples a unique laser-scanning device \cite{Scott_2002} with the
mass analyzer operating commercial Finnigan FT/MS software
(Bremen, Germany). It is capable of acquiring mass spectral data
from numerous locations on a sample while tracking the
x,y-positions. The positioning of the laser-scanning device and
acquisition of mass spectral data has been fully automated
\cite{McJ_JALA2002}. The I$^{2}$LD-FTMS generates a plethora of
data as up to 3600 files per hour can be acquired. Manual analysis
of this data would be a daunting task; therefore, we have
developed a data classifying agent to analyze the data and produce
a classification map of the sample.

Automation of mass spectra interpretation has been reported for
peptides and proteins \cite{Horn}, pharmaceuticals
\cite{Korfmacher}, and glycerolipids \cite{Kurvinen}. Some
researchers have applied all data points from a mass spectrum as
inputs to a neural network with an output for each classification
\cite{Wilkins96}. This method works well for complex spectra where
the number of inputs cannot be reduced. More complicated solution
surfaces lead to more computation requirements and less
transparency in the decision process. Training methods for the
neural networks have been applied \cite{Wilkins96,Ingram99}.
Others have defined branching decision trees to implement expert
systems \cite{Georgakopoulos}. Still others have defined typical
relative abundance of a set of {\em key} ions as a vector for each
class \cite{Ingram99}. They then chose the class with the minimum
Euclidean distance to a given sample spectrum.

%\pubidadjcol
Manual basalt mineral phase classification using mass spectra is
accomplished by analyzing the relative peak abundances versus
mass-to-charge (m/z) ratios. Traditionally, an analyst builds a
repertoire of spectral characteristics by inspecting spectra from
known homogeneous mineral types. Assignment of spectra from
unknown or heterogeneous samples is then accomplished by
comparison with the reference spectra of the known mineral types.
In the case of basalt phases, it was noticed that the relative
magnitude of specific key peaks to each other are the primary cues
for classifying the data. This handful of peaks corresponds to the
ions whose relative abundance determines the appropriate mineral
classification.

To classify basalt, Ingram et al. \cite{Ingram99} used a neural
network based vector quantization to group spectra and then assign
a classification to each group, providing a way to find
classification with a priori knowledge of only the significant
ions.  However, the Euclidean distance used in this method can
result in incorrect assignment of class due to the arbitrary
assignment of the average abundance of an ion that is {\em
unimportant} to a particular classification. An abundance for such
an ion could contribute to the Euclidean distance to the center of
an incorrect class being shorter than to the center of the
appropriate classification. Our method is similar in selecting a
small group of ions as inputs to classify the spectra. However, a
fuzzy logic \cite{Zadeh_65,CC_Lee_1,CC_Lee_2} membership function
approach is used in place of a Euclidean distance, which allows
full membership to be assigned over a range of relative abundance
rather than specifying a single exact value and also provides for
exclusion of ions from specific classifications (i.e. a logical
{\em don't care}).

Section \ref{fuzzy_thresh} describes the inference engine
developed to classify mass spectra and is illustrated for mineral
types found in basalt samples. The rules for the inference process
were derived from analyzing the process that a human analyst used
in classifying the mass spectra. The inference process was
distilled to a concise set of rules that could be implemented with
fuzzy logic. Subsequently, a method for building statistics, which
assists in determining appropriate rules, was developed. This
method, which also allows the identification of other key ions or
subclassifications and may be a basis for future work on
statistical classification methods, is described in Section
\ref{PeakTool}. In Section \ref{map}, a useful method for
classifying a mass spectrum, which is classified as unknown, based
on the membership values in mineral sets and the classification of
the location's neighbors is discussed. Finally, results and
conclusions are presented.Comparison of this fuzzy method to principle
component analysis and K-means clustering has been reported in Yan, et.al.
\cite{Yan_06}.

%\fedsection{Fractal Dimensions} \label{frac_sec}
\section{Classification of Spectra with Fuzzy Thresholds} \label{fuzzy_thresh}
A mineral phase in a basalt sample can be identified by its
chemical composition \cite{MineralBook}. In particular, the
relative abundance of particular ions give the signature for a
specific mineral type. The laser desorption process lifts both
ions of elements and molecules from the surface of the rock
sample. These ions are trapped in the Fourier transform mass
spectrometer cell where they are excited with a chirped radio
frequency field driven across a band from $50 hz \rightarrow 4Mhz$
at a sweep rate of $3500 hz/\mu S$. A Fourier transform (FT) is
applied to the digitized signal received from the sense plates,
which are orthogonal to the excite plates. The frequency scale of
the FT has a one-to-one correspondence with the m/z of the ion
whose excitation induced the frequency in the sense plates.

A natural fit for fuzzy logic was found when analytical chemist
described the decision for classification as: ``The basalt phase
Augite has significant abundance of iron, large abundance of calcium
and little or no titanium.'' The linguistic description, void of
explicit thresholds, provided a direct path to fuzzy logic expression
of classification rules.

Another reason for using fuzzy membership functions in FTMS mineral
identification is the magnitude of a particular peak is
proportional to the abundance of ions with the corresponding m/z.
However, for at least a couple of reasons, precise abundance
results are not expected. For one, the efficiency of the laser
desorption/ionization process varies for different elements and is
also influenced by the matrix that entrains the element;
therefore, the ion abundance is not strictly proportional to the
elemental percent composition expected for a given mineral.
Secondly, the finite spot size of the laser can desorb multiple
mineral types at a single sample location. A fuzzy logic based
decision can interpolate, with some success, between mineral types
when ions from various mineral types are ``blended''.

\subsection{Fuzzification}

Mass-to-charge peaks that affect the expert classification are
converted to a fuzzy logic level based on their relative
abundance. The truth level for a particular ion in a specific
classification can be Boolean by setting a threshold: when the
abundance is above the appropriate side of the threshold, the
truth level is high. It is convenient to allow for a {\it gray}
area with a piece-wise linear function mapping the abundance into
a $[0,1]$ range, so that an abundance just outside the threshold
can have a graduated approach in its affect on the classification.

A given mineral type will require that several ions be present or
absent in a range of relative abundances. The requirement for a
specific ion is encoded in a fuzzy membership function
$\mu_{\gamma,\chi}(A)$ where $\gamma$ is the mineral
classification, $\chi$ is a specific ion denoted as a chemical
symbol or the m/z of the ion, and $A$ is the mass spectrum of the
sample being classified. The spectrum, $A$, can be represented as
the relative abundance as a function of m/z, $a(\phi)$, where
$\phi$ is the m/z.  The first step in finding
$\mu_{\gamma,\chi}(A)$ is locating the maximum abundance, $p$,
within the error bound, $\epsilon$, of $\chi$:
\begin{equation}
p(A,\chi) = \max_{\phi = \chi - \epsilon}^{\chi +
\epsilon}(a(\phi)).
\end{equation}
The error bound is required because of uncertainty due not only to
drift in magnetic field strength between calibrations but to an
electric space-charge effect depending on the number of ions
desorbed \cite{Marshall} (pages 244-245).

{
  \renewcommand{\baselinestretch}{1.0}
  \begin{figure}
    \centerline{
                \includegraphics[width=3.5in]{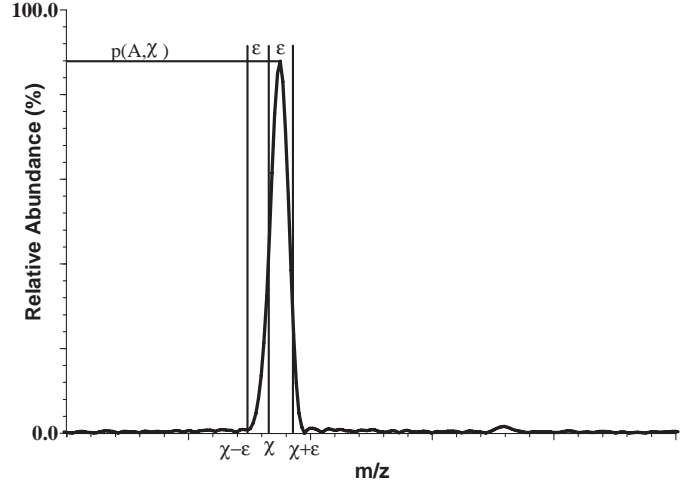}
               }
    \caption{Illustration of maximum abundance within $\epsilon$ of $\chi$.\vspace{.25in}
    }
    \label{fig_peak}
  \end{figure}
}

The membership value then becomes a function mapping $p$ onto a
fuzzy logical level, $[0,1]$. The linguistic expression for
whether an ion's abundance is appropriate for a particular mineral
composition can take a form similar to: ``the relative abundance
of iron should be small'' or ``the relative abundance of calcium
(Ca) should be high.'' The level at which the logic level is
$100\%$ true or false is a judgement call, which can better be
performed by allowing interpolation. Experts could conceivably
compromise on levels of abundance that constitute an absolute true
or false and allow a function to interpolate between those levels.
For lack of apparent need for a more complex interpolation, we
chose to implement a piece-wise linear function. In general we
could have medium relative abundance functions; but, in practice,
we have only found need to define functions for relative high or
low abundances. The low (not) abundance function can be formed as
the negative of a high abundance function. So, $\mu$ takes one of
the two forms, shown in Fig. \ref{fig_membership}:

\begin{equation}
\mu_{\gamma,\chi}(p) = \left\{ \begin{array}{ll}
            0 & p<l \\
            \frac{ p-l }{ h-l }  & l \leq p < h \\
            1 & p \geq h
            \end{array} \right.
\end{equation}

for required high abundance, or for low abundance

\begin{eqnarray}
\mu_{\gamma,\sim\chi}(p) & = & 1 - \mu_{\gamma,\chi}(p) \nonumber \\
               & = & \left\{ \begin{array}{ll}
            1 & p<l \\
            \frac{ h-p }{ h-l }  & l \leq p < h \\
            0 & p \geq h
            \end{array} \right.,
\end{eqnarray}
where $l$ is low hard logic threshold and $h$ is the high logic
threshold, $l < h$, and the symbol `$\sim$' indicates the negation
(i.e. NOT $\chi$). The primary method for choosing $l$ and $h$ has
an expert operator look at many spectra taken from various
locations on a homogeneous ``known.'' From these observations
thresholds can be set and the size of the ``gray'' area chosen.

{
  \renewcommand{\baselinestretch}{1.0}
  \begin{figure}
    \centerline{
                \includegraphics[width=3.5in]{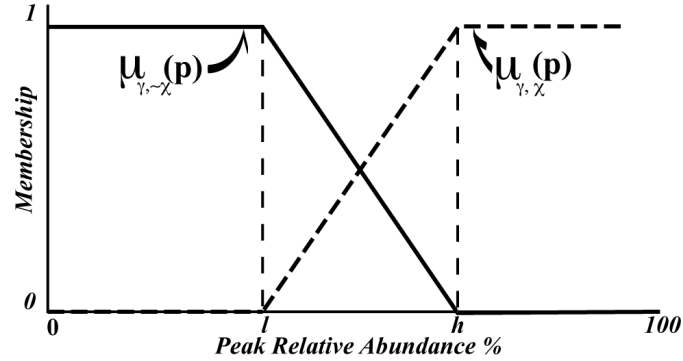}
               }
    \caption{Examples of membership function for high and low abundance requirements.\vspace{.25in}
    }
    \label{fig_membership}
  \end{figure}
}

\subsection{Inference Rules}
Inference rules are based on logical expressions of requirements.
The fuzzy membership functions determine the truth value of each
term in the expression. Logical operations are defined using the
logic levels from the fuzzy membership values of an ion with
respect to the mineral class. The ``and'' ($\logand$) function is
implemented with the product operation ($A \logand B \rightarrow
\mu_A \mu_B$), the ``or'' ($\logor$) function with the equivalent
($A \logor B \rightarrow \mu_A + \mu_B - \mu_A \mu_B$), and
``not'' ($\sim$) as negation, $1-\mu$.  As an example, our rule
for augite (AGT) has requirements on the ions of iron (Fe),
titanium (Ti) and Ca. The expression for membership in the class
$\mbox{AGT}$ is the product of the result of the membership
functions, $\mu_{{AGT},Fe}$, $\mu_{{AGT},\sim Ti}$, and
$\mu_{{AGT},Ca}$ and can be written compactly as:
%\begin{equation}
%\mu_{{AGT}} = \mu_{{AGT},Fe} \otimes \mu_{{AGT},Ti} \otimes
%\mu_{{AGT},Ca}
%\end{equation}
%which for our purpose is the same as
\begin{equation}
%\mu_{{AGT}} = \min(\mu_{{AGT},Fe}; \mu_{{AGT},Ti};
%\mu_{{AGT},Ca}).
\mu_{AGT}(A) = \Logand_{\chi \in}^{\{Fe,\sim Ti,Ca\}}
\hspace{-12pt}\mu_{{AGT},\chi}(A).
\end{equation}

The benefit of using the product operation over minimums and
maximums is the additive affect in the ``or'' terms. In
particluar, ilmenite is defined as having one or more of the
metals, Fe, magnesium (Mg), or manganese (Mn), above a certain
accumulative abundance. This allows the membership function to be
constructed such that appropriate combination of the metals will
give a sufficiently high membership value. The product
implementation for ``and'' terms also discounts membership of
multiple terms below full membership, providing a more
conservative approach (e.g. two terms of membership value $0.9$
``and''-ed together discounts the resulting membership value to
$0.81$ instead of the minimum value $0.9$).

\subsection{Hard Classification}

Although leaving the result of the classification as the value of
the fuzzy membership in each the mineral classes is useful in some
applications, our current application requires discrete
classifications. Defuzzification is accomplished by finding the
mineral class with the maximum membership value. A threshold,
$\nu$, is set at a minimum required membership. If the spectra
does not have a maximum membership in one of the classes greater
than $\nu$ then it is classified as unknown, UNK. The hard
classification, $\Gamma$ is:
\begin{equation}\label{hard_class}
\Gamma(A) = \left\{ \begin{array}{ll}
    (\gamma | \mu_\gamma = \displaystyle\max_{\chi}\mu_\chi) & \mbox{for} \displaystyle\max_{\chi}\mu_\chi \geq \nu\\
    \mbox{UNK} & \mbox{for} \displaystyle\max_{\chi}\mu_\chi < \nu
 \end{array} \right.
\end{equation}
As a side note, the membership value in the set of UNK can be
defined as the negation of the maximum membership value:
$\mu_{\mbox{UNK}} = 1 - \max_{\chi}\mu_\chi$.

Unknown spectra are left for the instrument operator to classify
or remain as `unknown.' An unknown spectra may occur when the
desorption laser spot covers more than one class of mineral or
there is a problem with the desorption process or acquired signal.
An automated method for assigning a value based on the membership
values and the neighboring spots in the sample is presented in
Section \ref{map}.

One item of note, some basalt samples from the INL site have the
characteristic of an unusually high potassium abundance not noted
in the mineral literature \cite{MineralBook}. This has been
reported in other mass spectrometry literature for analysis of INL
basalt \cite{Ingram99}. To adjust for the high potassium
abundance, spectra used by the inference engine are preprocessed
to re-scale based on the highest peak excluding potassium. The
software implementing the system allows particular ions to be
rescaled in this way when required.

%Section \ref{results} contains Table \ref{tab:Basalt_inf} showing
%a rule base typically used for spectra from INL's instrument.

\section{Ensemble Statistics for Generating and Refining Rulebases} \label{PeakTool}
In this section, a spectra statistical method is presented for determining or
refining rules given a set of data. The
method acts on a set of spectra that have been classified as the
same type of mineral by the rule base, are from a known
homogeneous sample, or from a diverse ensemble.
Several statistics are accumulated across all sample data about the
peaks in the spectra: sum, sum of squares, minimum and maximum of
abundances, and count of non-zero abundances. The size of bins
used to sort peaks into is determined by the mass accuracy of the
FTMS.

Given a set of spectra, $F$, which have the same classification,
composed of $\{A_1, ..., A_n\}$, we want to look for similarities
in the mass spectra. A data base containing m/z, number of spectra
with a peak of the m/z, the total relative abundance, and the sum
of the square of the abundance is generated. A list of local maxima
or ``peak list'' of a spectrum can be used to represent
$A$, $\bar{a} = \{(\phi[1], \bar{a}[1]), ... ,(\phi[n], \bar{a}[n])\}$,
where $n$ is the number of local maxima in $a(\phi)$ and the list
is sorted on $\phi$. If two or more peaks have $\phi$ within $\epsilon$
of each other, we form a consolidated list ($\hat{a}$) of only
the maximum abundance of these peaks.

A statistical data base element, $S[i]$, is created for each bin
of $\phi$ within $\epsilon$ of each other. The fields of the data
base will be noted with $S[i].field$. The first element is the m/z
of the bin, $S[i].\phi$. We chose to calculate this as the mean of
the $\phi$ in spectra with a peak in the $i^{\mbox{th}}$ bin,
though there are many ways (weight average, $\phi$ of maximum
abundance, etc.) which could be used. The total count of spectra
containing a non-zero abundance peak within $\epsilon$ of $\phi$
is tabulated in $S[i].c$. The sum, sum of squares, max, and min of
the abundance for the peaks with spectra fitting the same
requirement are accumulated in $S[i].a_{tot}$, $S[i].a_{tot}^2$,
$S[i].a_{max}$, and $S[i].a_{min}$, respectively. These are typical
statistics for determining mean, standard deviation, and count for a given
m/z peak.

Additional key ions that positively discriminate one class versus
the remainder of the classifications can be determined using the
statistics of spectra from that class versus the statistics of the
ensemble of the diverse spectra from the other classes. By
selecting the $S[i]$ with $S[i].c$ equal to the total count of
spectra classified in the class, candidates for unique positive
cues for the class are obtained. When the mean of the abundances
$S[i].a_{tot}/S[i].c$ is compared to the mean of the abundances
for data in all classifications, a ratio significantly larger than
$1$ is motivation to investigate the peak as a possible key ion.
Alternatively, a result of a small value for $S[i].a_{tot}$
divided by the total count of spectra in the classification,
including the zero abundance spectra in the average, can be
compared to the mean of the whole of the data. If the ratio of the
class mean to the ensemble the peak may also provide a
distinguishing characteristic of a low abundance compared to the
entire set. An example is shown in Section \ref{results}.

This set of statistics can be utilized to locate distinct peaks to
this classification, if a particular $S[i]$ is unique to the
classification $F$. It can also be used to identify
sub-classifications or contaminants, if a subset of $F$ all
contain unique peaks compared to the rest of $F$. A bifurcation of
the classification into subclassifications may be indicated by a
count of abundances less than the total count of the spectra in
the set. A further extension of this concept would be to look for
separation(s) in the abundances for a given m/z into two or more
clusters.

Finally if the spectra are from a homogeneous sample of material,
which does not have a set of inference rules, a rule base can be
derived from these statistics using key peaks. While not currently
automated, the tools outlined in this section provide methods that
are used to create and refine rules based on accumulated
statistics.

\section{Spatially Based Classification of Indeterminate Samples} \label{map}
Occasionally, a spectrum will be a combination of minerals types
or be overwhelmed by an anomalous ion, such as potassium, or of
low magnitude because of degraded vacuum. For cases where there
are sporadic ``unknown'' spots, a method for filling in the
unknowns based on the surrounding spots is desirable. Such a
method is described in this section.

The membership value of a particular spot for a class, $\gamma$,
is denoted by, $\mu_{\gamma}[i]$, where $i$ is an index for the
spot. The index, $i$, maps to a Cartesian coordinate of the spot.
Typical spot patterns are either rectangular grids or hexagonal
(closest pack) grids, see Fig. \ref{fig_neighbors} . For this
method we identify the neighbors of the unknown spot and modify
the membership value of the unknown spot based on the the
membership values of the surrounding spots:
\begin{equation}
{\bar{\mu}}_{\gamma}[i] = \mu_{\gamma}[i] + \displaystyle
\sum_j^{\mbox{neighbors}}\frac{\mu_{\gamma}[j]}{N},
\end{equation}
where $N$ is the number of neighboring spots. The membership value
of spot $i$ is added to the average of the spots near it. The hard
classification is made by finding the maximum membership value:
\begin{equation}\label{hard_class_map}
 \Gamma[i] = \left\{ \begin{array}{ll}
    (\gamma | \mu_\gamma[i] = \displaystyle\max_{\chi}\mu_\chi[i]) & \mbox{for} \displaystyle\max_{\chi}\mu_\chi[i] \geq \nu\\
    (\gamma | {\bar{\mu}}_\gamma[i] = \displaystyle\max_{\chi}{\bar{\mu}}_\chi[i]) & \mbox{for} \displaystyle\max_{\chi}{\mu}_\chi[i] < \nu

 \end{array} \right. .
\end{equation}
Fig. \ref{fig_neighbors} shows the neighboring spots which would
be selected. If the spot is at an edge or corner of the data set
$N$ would be reduced appropriately for missing spots. {
  \renewcommand{\baselinestretch}{1.0}
  \begin{figure}
    \centerline{
                \includegraphics[width=2.5in]{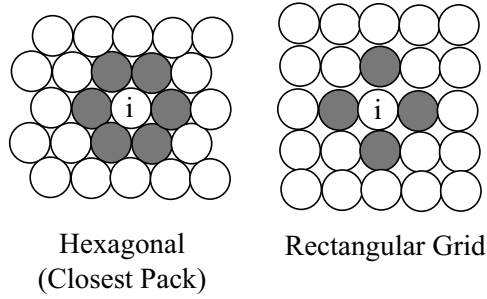}
               }
    \caption{Neighbors used for classifying unknown are shown as gray for the two typical spot patterns.\vspace{.25in}
    }
    \label{fig_neighbors}
  \end{figure}
}

\section{Results and Discussion} \label{results}
The methods described in the previous sections have been applied
to basalt samples. This section will describe the resulting rule
base and the steps taken to arrive at the current solution.
Finally, a set of results from a sample of basalt is given with
discussion on the result.

The rules were developed in a progression: (1) FTMS operator
observed spectra from homogeneous samples of the expected mineral
phases; (2) additions and modifications to the rules where made
based on survey of mineral reference literature
\cite{MineralBook}; (3) the tool for grouping statistics was
developed and used to identify additional ions and shape
membership functions; and (4) a method for making a good guess at
spots that were {\em unknown} to the inference engine was
developed for data presentation requiring a complete map.

The use of fuzzy logic was prompted by the realization that the
FTMS operator used phrases such as, ``Ilmenite has a high
abundance of Ti and Fe compared to other mineral phases.'' Based
on tabulated chemical composition data \cite{MineralBook} from
various samples of the mineral phases of interest from around the
world, the rules were refined. For example, the reference
literature made clear that plagioclase should contain a
significant abundance of Al. Aluminum was initially overlooked
because its relative abundance is lower than that of the primary
cue of Ca from the plagioclase samples. In fact, Ca is a {\em don't
care} ion because some varieties of plagioclase do not contain Ca.
This emphasizes that a human can be fooled without the appropriate
visualization tool.

\begin{figure}
    \centerline{
                \includegraphics[width=3.5in]{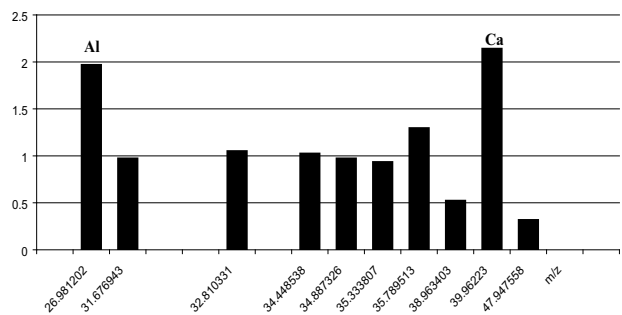}
               }
    \caption{Average abundance of peaks in plagioclase samples spectra plagioclase
     relative to average abundance an ensemble of spectra from various ``knowns.''
     Only ions which were contained in all plagioclase samples are displayed in
     the histogram. Note only Al and Ca are significantly above the
     average of the ensemble and Ca is known to be
     absent in some plagioclase from other geological areas.\vspace{.25in}
    }
    \label{fig_histo}
  \end{figure}

Although the appropriate use of compiled reference data should not
be undervalued, the tool for accumulating statistics also showed
that Al was an important indicator in plagioclase. Fig.
\ref{fig_histo} displays the ratio of the average abundance of
ions occurring in a homogeneous sample of plagioclase to the
average abundance of the ions in an ensemble of all mineral phase
samples tested. Only the ions which were present in all of the
plagioclase spectra are shown. The only ions whose average
abundance in plagioclase is significantly above the ensemble
average are Al and Ca. Al is an important constituent that was
initially overlooked because the relative magnitude of the
abundance was small compared to the other ions. If one has
confidence in the known samples, the visualization tool can be
used to find the unique cues between types that will allow for
classification rules to be generated.
  \renewcommand{\baselinestretch}{1.0}

Finally, there are circumstances where the spectrum from a spot
are not identifiable. For example, in the case of degraded vacuum
with the FTMS, the signals are too weak to generate a spectra
above the noise level threshold set for the system. A more
typical example is when a spot is too large to desorb only one
mineral type. For applications where we need to classify all
spots regardless of confidence in the data, we argue that
proximity of like minerals should influence the decision. To this
end, the method in Section \ref{map} was created.

The result of this process has development of the rules listed in
Table \ref{tab:Basalt_inf} for classification of the four basic
types of minerals found in basalt in the Snake River vadose zone.
Those types include: olivine (OLV), augite (AGT), ilmenite (ILM),
and plagioclase (PLG).

\begin{table}
 \caption{Definition of inference engine
for classification of mineral types in basalt.
\label{tab:Basalt_inf} }
%\begin{center}
\begin{tabular}{l|c|c}
\hline \hline \multicolumn{3}{c}{Ilmenite (ILM)} \\
\multicolumn{3}{c}{ $ \mu_{\mbox{ILM}}=\displaystyle \Logand_{\chi \in}^{\{Fe,Ti,\sim Al\}} \mu_{\mbox{ILM},\chi}$} \\
\hline \hline
$\chi$ (Chem/mz) & \hspace{.4in}$l$ \hspace{.4in} & $h$ \\
\hline
$\sim Al/26.982$ & $0.5$ & $15$\\
$Ti/47.95$ & $1$ & $17$\\
$Fe/55.954$ & $1$ & $40$\\
\hline \hline  \multicolumn{3}{c}{Augite (AGT)} \\
\multicolumn{3}{c}{ $ \mu_{\mbox{AGT}}=\displaystyle \Logand_{\chi \in}^{\{Fe,\sim Ti,Ca\}} \mu_{\mbox{AGT}, \chi} $} \\
\hline \hline
$\chi$ (Chem/mz) & $l$ & $h$\\
\hline
$Ca/39.95$ & $50$ & $80$\\
$\sim Ti/47.95$ & $1$ & $17$\\
$Fe/55.954$ & $1$ & $30$\\
\hline \hline  \multicolumn{3}{c}{Plagioclase (PLG)} \\
\multicolumn{3}{c}{ $ \mu_{\mbox{PLG}}= \displaystyle \Logand_{\chi \in}^{\{Al,\sim Fe,\sim Ti\}} \mu_{\mbox{PLG}, \chi}$} \\
\hline \hline
$\chi$ (Chem/mz) & $l$ & $h$\\
\hline
$Al/26.982$ & $0.5$ & $15$\\
$\sim Ti/47.95$ & $1$ & $17$\\
$\sim Fe/55.954$ & $10$ & $40$\\
\hline \hline  \multicolumn{3}{c}{Olivine (OLV)} \\
\multicolumn{1}{r}{$\mu_{\mbox{OLV}}=$} & \multicolumn{2}{l}{ $ (\mu_{\mbox{OLV}, Mg} \logor \mu_{\mbox{OLV}, Mn} \logor \mu_{\mbox{OLV}, Fe})$} \\
\multicolumn{1}{r}{} & \multicolumn{2}{l}{$\logand \sim\mu_{\mbox{OLV}, Ti}  \logand \sim\mu_{\mbox{OLV}, Al}$} \\
 \hline
\hline
$\chi$ (Chem/mz) & $l$ & $h$\\
\hline
$Mg/24.312$ & 1 & 50\\
$\sim Al/26.982$ & $0.5$ & $15$\\
$\sim Ti/47.95$ & $1$ & $17$ \\
$Mn/54.938$ & $10$ & $40$ \\
$Fe/55.954$ & $10$ & $40$ \\

\hline \hline
\end{tabular}
%\end{center}
\end{table}

%  \renewcommand{\baselinestretch}{1.0}
%  \begin{figure}
%    \centerline{
%                \includegraphics[width=3in]{map_wnn.eps}
%               }
%    \caption{Mapping of an area of a sample of basalt from the
%              Snake River vadose zone. Four graphs, b)-e), show the membership
%              value of the spots.  The ``hard'' classification
%              map, a), is shown
%              without applying the technique in Section \ref{map}. Result after
%          applying near neighbor contributions is shown in f). Spots
%              spaced $30\mu m$ apart (center to center).\vspace{.25in}
%    }
%    \label{fig_map}
%  \end{figure}

  \renewcommand{\baselinestretch}{1.0}
  \begin{figure}
    \centerline{
                \includegraphics[width=3in]{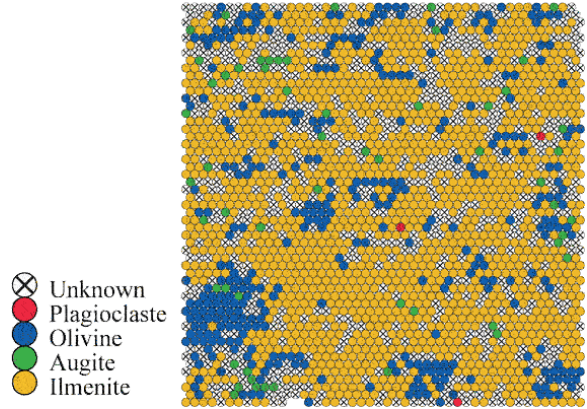}
               }
    \caption{Classification mapping of an area of a sample of basalt from the
              Snake River vadose zone. Four graphs, b)-e), show the membership
              value of the spots.  This is a {\em hard} classification
              map without applying the technique in Section \ref{map}.  Spots
              are spaced $30\mu m$ apart (center to center).\vspace{.25in}
    }
    \label{fig_map}
  \end{figure}

\begin{figure}
    \centerline{
                \includegraphics[width=3in]{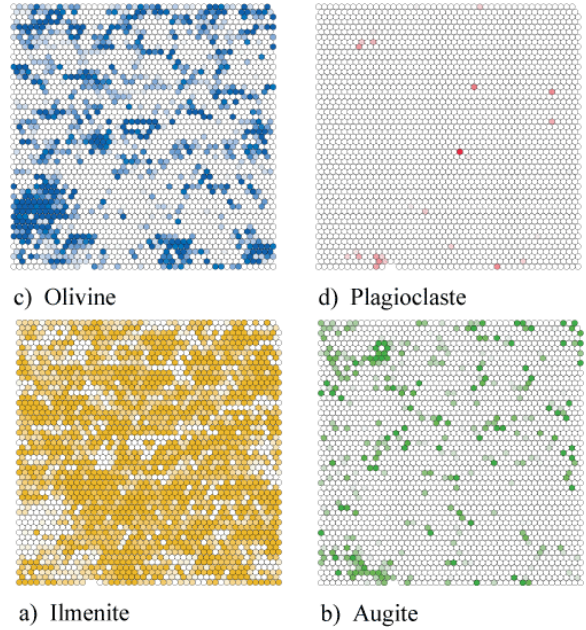}
               }
    \caption{Fuzzy membership value for each classification.\vspace{.25in}
    }
    \label{fig_map_conf}
  \end{figure}

\begin{figure}
    \centerline{
                \includegraphics[width=3in]{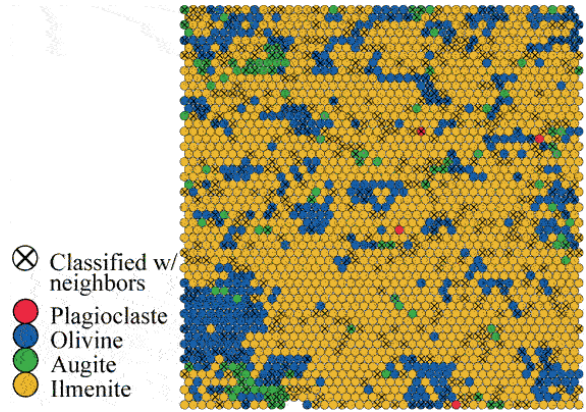}
               }
    \caption{Classification mapping after applying near neighbors to unknowns.\vspace{.25in}
    }
    \label{fig_map_neigh}
  \end{figure}

The map in Fig. \ref{fig_map} shows an image of a section of
basalt from the INL site. Without applying the technique described
in Section \ref{map} and choosing $\nu = 0.5$, the hard
classification in Fig. \ref{fig_map} shows some unknown spots.
Membership values in individual classification is shown in Fig.
\ref{fig_map_conf}. Fig. \ref{fig_map_neigh}. shows the result of
using the neighboring spots to help classify the unknowns. It is
possible that the spot size used in this example was too large to
be focused on individual mineral phases for some of the spots,
where the two adjacent phases have a pair of mutually exclusive
ion requirements and both are assigned a low membership value,
resulting in the unknown classification. For example, the spot
could contain both ilmenite and plagioclase, so the appearance of
significant amounts of Ti and Al eliminate both classes from
consideration. In another example, a high membership value of two
classes indicates an overlap even though the hard classification
chooses the highest level. Both of these cases would be resolved,
except on the exact borders of mineral phases, by reducing the
laser desorption spot size. However, there may always be cases
where some of the mineral microcrystals are too small for the
smallest laser spot size used. Therefore, using the nearest
neighbor protocol will allow the unknown spots to be assigned and
the membership value can be referenced to find a overlap of
classes to a spot. Alternatively the hard classification can be
adjusted to assign unknown to the case where we have multiple high
membership value.

  \renewcommand{\baselinestretch}{1.0}
  \renewcommand{\baselinestretch}{1.0}

\section{Conclusion and Future Directions} \label{conclusion}
The fuzzy logic inference engine described in the paper has been
shown to accurately classify minerals in basalt. The automated
classification system the inference engine provides is required to
process the large amounts of data produced by an imaging FTMS. It
has some advantages over other schemes used in the field.
Specifically, it is simpler and more transparent in operation than
neural networks. Also, the inference engine has flexibility to
implement {\em don't care} statements for the presence of
specified ions, because Another important feature is the ability
to leave a {\em don't care} ion out of the logic expression, which
avoids the problem of using an Euclidean measure. The output of
the fuzzy logic inference engine provides a confidence level that
can allow a human operator's attention be drawn to the
questionable and possibly interesting data in a large image.
Additionally, a statistical visualization tool has been created to
assist in developing rules for classification. Finally, a method
for assigning indeterminate spots was described and illustrated.

In the future, rule bases for different types of materials will be
created. For example, classification of soils will also be
examined with the system. Additionally, the presence and type of
microorganisms residing on basalt will be incorporated into the
rule base. The input to the rule base will also be extended to the
optical spectra that are also acquired as part of the overall LOCI
instrument. Additionally automation of the methods discussed for
generating and refining of rule bases will be investigated.

\section{Acknowledgements}
We thank Paul L. Tremblay for editorial suggestions to this paper
and for being instrumental in the design and construction of the
imaging FTMS. This work was supported by the U.S. Department of
Energy under DOE/NE Idaho Operations Office Contract
DE-AC07-05ID14517.
\section{List of Symbols} \label{symbols}
For the readers convenience, the following is a list of symbols
and variable notations used throughout the paper.

{\renewcommand{\arraystretch}{1.3}
\begin{tabular}{lcl}
$A$ \symbolsep
{FTMS spectrum to be analyzed.} \\

$\gamma$ \symbolsep
{variable to represent a mineral classification}\\

$\chi$
\symbolsep
{variable for a particular ion, indicated by either the chemical symbol or the m/z of the ion.}\\

$\phi$
\symbolsep
{variable for m/z.}\\

$a(\phi)$
\symbolsep
{a function representing $A$ as the abundance as a function of m/z.}\\

$\hat{a}$ \symbolsep
{peak list representation of $A$ as ordered pairs $(\bar{a}, \phi)$ of local maxima of $a(\phi)$.}\\

$p(A,\chi)$ \symbolsep
{peak abundance of spectrum $A$ within $\epsilon$ of m/z for ion $\chi$}\\

$\epsilon$ \symbolsep
{error bound around the m/z for an ion $\chi$ for finding peak abundance.}\\

$\mu_{\gamma}(A)$ \symbolsep {membership value for $A$ in the
mineral classification, $\gamma$.}\\

$\mu_{\gamma,\chi}(A)$ \symbolsep
{membership value for a FTMS spectrum $A$ for mineral classification $\gamma$ and ion $\chi$.}\\

$\mu_{\gamma,\chi}(p)$ \symbolsep
{membership value requiring a high abundance for the peak, $p$, for mineral classification $\gamma$ and ion $\chi$.}\\

$\mu_{\gamma,\sim\chi}(p)$ \symbolsep
{membership value requiring a low abundance for the peak, $p$, for mineral classification $\gamma$ and ion $\chi$ (equivalent to $1-\mu_{\gamma,\chi}(p)$).}\\

$\mu_{\gamma}[i]$ \symbolsep {membership value of the $i^{th}$ desorption spot in the class, $\gamma$.}\\

${\bar{\mu}}_{\gamma}[i]$ \symbolsep {membership value of the $i^{th}$ desorption spot in the class, $\gamma$, biased by its neighboring spots.}\\

$\logand$ \symbolsep {fuzzy logic ``and'' operator defined for
this paper as the product of the two operands, $\mu_A \mu_B$.}\\

%the tabular environment does not split at page or column breaks so
%this tabular end and begin must be inserted at the appropriate place
%depending on its location in the text.

$\logor$ \symbolsep {fuzzy logic ``or'' operator defined as the
logical inverse of the product of the two operands, $\mu_A + \mu_B - \mu_A\mu_B$.}\\

\end{tabular}

\begin{tabular}{lcl}

$\nu$ \symbolsep {threshold value setting the minimum membership
value to determine the ``hard'' classification}\\

$\Gamma(A)$ \symbolsep {Defuzzification function returning the
``hard'' classification or ``unknown.''}\\
$\Gamma[i]$ \symbolsep {Defuzzification function returning the
``hard'' classification for the $i^{th}$ desorption spot.}\\

$S[i].\phi$ \symbolsep {Center m/z of the $i^{\mbox{th}}$ bin}\\
$S[i].a_{tot}$ \symbolsep {Total abundance from all
spectra---used to compute the mean}\\

$S[i].a_{tot}^2$ \symbolsep {Sum of the square of abundance---for calculation of standard deviation}\\
$S[i].a_{max}$ \symbolsep {Maximum abundance for $i^{\mbox{th}}$
bin}\\

$S[i].a_{min}$ \symbolsep {Minimum abundance}\\
$S[i].c$ \symbolsep {Count of spectra with a non-zero peak in the
$i^{\mbox{th}}$ bin.}\\

%below is the example of using the symbol list separator
%$ $ \symbolsep {}\\
\end{tabular}
}

\newpage
%\doublespace
\bibliography{IEEEabrv,fuzzy,chem,mcj}
\bibliographystyle{klunamed}
%\begin{thebibliography}{99}
%  \input{references.tex}
%\end{thebibliography}
%\input{bios.tex}
%\input{bios.tex}
%figure headings for draft
%\newpage \pagestyle{empty}
%\section{Figure captions}
%\input{figure_headings.tex}
\end{article}
\end{document}